\UseRawInputEncoding

\documentclass[runningheads]{llncs}
\usepackage{graphicx}
\usepackage{tikz}
\usepackage{comment}
\usepackage{amsmath,amssymb} 
\usepackage{color}
\usepackage[T1]{fontenc}

\usepackage[utf8]{inputenc} 

\usepackage[main=english,romanian]{babel}
\usepackage{combelow}

\usepackage[accsupp]{axessibility}  

\usepackage{bbold}

\begin{document}
\pagestyle{headings}
\mainmatter
\def\ECCVSubNumber{14}  

\title{PVBM: A Python Vasculature Biomarker Toolbox Based On Retinal Blood Vessel Segmentation} 

\titlerunning{PVBM}
%
\author{Jonathan Fhima\inst{1,2} \and
Jan Van Eijgen\inst{3,4} \and
Ingeborg Stalmans\inst{3,4}\and Yevgeniy Men\inst{1,5} \and Moti Freiman \inst{1} \and Joachim A. Behar \inst{1}}
\authorrunning{Fhima J. et al.}
%
\institute{Faculty of Biomedical Engineering, Technion-IIT, Haifa, Israel \and
 Department of Applied Mathematics Technion–IIT, Haifa, Israel\and
 Research Group Ophthalmology, Department of Neurosciences, KU Leuven, Belgium \and
 Department of Ophthalmology, University Hospitals UZ Leuven, Belgium \and
 The Andrew and Erna Viterbi Faculty of Electrical \& Computer Engineering, Technion–IIT, Haifa, Israel}
\maketitle

\begin{abstract}
\textbf{Introduction: } Blood vessels can be non-invasively visualized from a digital fundus image (DFI). Several studies have shown an association between cardiovascular risk and vascular features obtained from DFI. Recent advances in computer vision and image segmentation enable automatising DFI blood vessel segmentation. There is a need for a resource that can automatically compute digital vasculature biomarkers (VBM) from these segmented DFI.  \textbf{Methods:} In this paper, we introduce a Python Vasculature BioMarker toolbox, denoted PVBM. A total of 11 VBMs were implemented. In particular, we introduce new algorithmic methods to estimate tortuosity and branching angles. Using PVBM, and as a proof of usability, we analyze geometric vascular differences between glaucomatous patients and healthy controls.  \textbf{Results:} We built a fully automated vasculature biomarker toolbox based on DFI segmentations and provided a proof of usability to characterize the vascular changes in glaucoma. For arterioles and venules, all biomarkers were significant and lower in glaucoma patients compared to healthy controls except for tortuosity, venular singularity length and venular branching angles.
 \textbf{Conclusion:} We have automated the computation of 11 VBMs from retinal blood vessel segmentation. The PVBM toolbox is made open source under a GNU GPL 3 license and is available on physiozoo.com (following publication).

\end{abstract}

\section{Introduction}
According to the National Center for Health Statistics, cardiovascular diseases (CVD), including coronary heart disease and stroke, are the most common cause of death in the USA \cite{Murphy2021Mortality2020}. 
Since the beginning of the 20th century, researchers have shown that retinal microvascular abnormalities can be used as biomarkers of CVD \cite{Gunn1892OphthalmoscopicTension,Gunn1898OphthalmoscopicDisease,Keith1939SomePrognosis}.
Retinal vasculature can be non-invasively assessed using DFIs, which can be easily obtained using a fundus camera. Consequently, retinal vascular features obtained from DFI may be used to characterize and analyze vascular health. In order to enable reproducible research, it is necessary to fully automate the computation of these biomarkers from the segmented vasculature. We developed a Python Vasculature Biomarker Toolbox (PVBM), based on DFI segmentations made by expert annotators from the University Hospitals Leuven in Belgium. PVBM enables a quantitative analysis of the vascular geometry thereof with broad application in retinal research. In this paper we illustrate the potential of PVBM by characterizing vascular changes in glaucoma patients.

\subsection{Prior works}
\subsubsection{Connection between retinal vasculature and cardiovascular health:}
\begin{figure*}[!tb]
    \centering
	\includegraphics[width=0.45\textwidth]{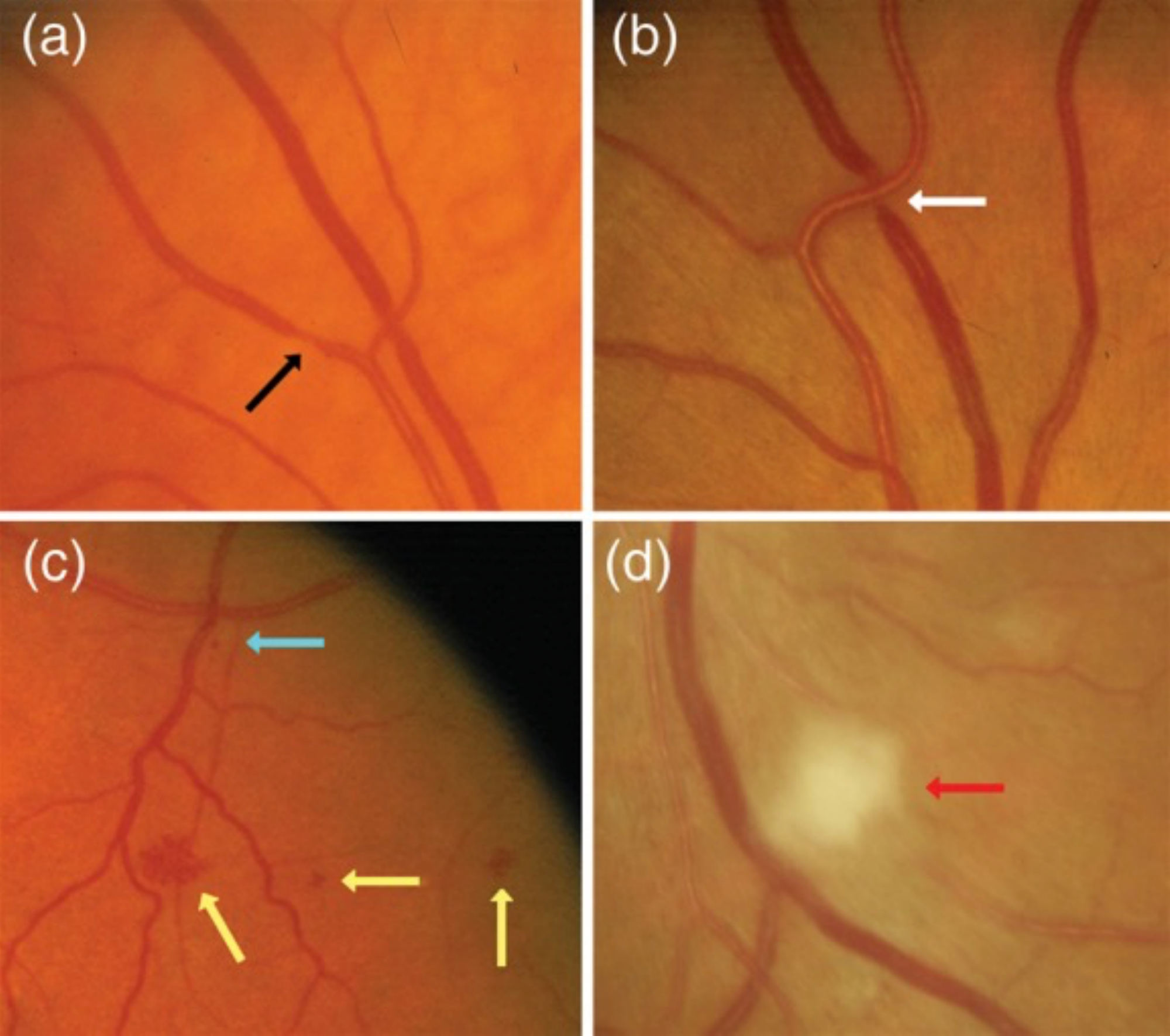}
	\caption{Examples of retinal vascular signs in patients with cardiovascular diseases. Reproduced with permission from Liew et al. \cite{Liew2011RetinalHeart}
	. Black arrow: focal arteriolar narrowing, White arrow: arterio-venous nicking, Yellow arrow: haemorrhage, Blue arrow: micro-aneurysm, Red arrow: cotton wool spot. }
	\label{fig:2}
\end{figure*}

As early as of beginning of the 20th Century research has been carried out to assess the relationship between the retinal vasculature and cardiovascular health. Marcus Gunn can be seen as one of the first to describe the relation between hypertension and retinal characteristics \cite{Gunn1892OphthalmoscopicTension,Gunn1898OphthalmoscopicDisease}, and is followed by the work of H.G Scheie \cite{Scheie1953EvaluationSclerosis} in 1953. In 1974, N.M Keith showed that hypertension and its mortality risk is reflected in the retinal vasculature \cite{Keith1939SomePrognosis} and in 1999 Sharrett et al. \cite{Sharrett1999RetinalStudy} added arterio-venous nicking and arteriole narrowing to the list of pathological findings. Examples of retinal microvascular abnormalities in hypertensive patients can be seen in Fig. \ref{fig:2}. Witt et al. \cite{Witt2006AbnormalitiesStroke} concluded that vessel tortuosity significantly distinguished between patients with ischemic heart disease and healthy controls. Over the years retinal vessel calibres were shown to change in hypertension \cite{BetzlerRetinalDiabetes}, obesity \cite{Hanssen2011Exercise-inducedObesity}, chronic kidney disease \cite{Mehta2015PhosphateStudy,Sabanayagam2009RetinalPopulation}, diabetes mellitus \cite{BetzlerRetinalDiabetes}, coronary artery disease \cite{Guo2020AssociationMeta-analysis} and glaucoma \cite{Kawasaki2013RetinalStudy}. Fractal dimensions of the retinal vascular tree are among the newest biomarkers to study cardiovascular risk. Monofractal dimension was shown to change with age, smoking behaviour \cite{Lemmens2021AgerelatedBehaviour}, blood pressure \cite{Liew2008ThePressure}, diabetic retinopathy \cite{Cheung2009QuantitativeAnalysis}, chronic kidney disease \cite{Sng2010FractalDisease}, stroke \cite{Doubal2010FractalStroke}, and coronary heart disease mortality \cite{Liew2011FractalMortality}, while Multifractal dimensions were found to be negatively associated with blood pressure and WHO/ISH cardiovascular risk score \cite{VanCraenendonck2021RetinalPopulation}. The established association between vascular biomarkers and CVD prompts the development of algorithmic solutions for automated computation thereof.

\subsubsection{Automated biomarker computation based on DFIs:}
Several attempts have been made to extract meaningful biomarkers to characterize cardiovascular health based on DFI vasculature. In 2000, Martinez-Perez et al. \cite{Martinez-Perez2000GeometricalImages} introduced a semi-automated algorithms capable of computing vasculature biomarkers (VBMs) such as vessel diameter, length, tortuosity, area and branching angles. In 2011, Perez-Rovira et al. \cite{Perez-Rovira2011VAMPIRE:REtina} created Vessel Assessment and Measurement Platform for Images of the REtina (VAMPIRE), a semi-automatic software that can extract the optic disk and compute vessel width, tortuosity, fractal dimension, and branching coefficient. RetinaCAD was developed in 2014 \cite{Dashtbozorg2014RetinaCADChanges}. This automated system is able to calculate Central Retinal Arteriolar Equivalent (CRAE), the Central Retinal Venular Equivalent (CRVE), and the Arteriolar-to-Venular Ratio (AVR). Lastly, many algorithmic approaches have been developed to estimate the blood vessel tortuosity  \cite{Hart1999MeasurementTortuosity,Grisan2008ATortuosity,Owen2009MeasuringProgram}. Last year, Provost et al. \cite{Provost2021DenserChildren} used the MONA REVA software which semi-automatically segments retinal blood vessels and measures tortuosity and fractal dimension in order to analyze the impact of their changes on children's behaviour.

\subsection{Research gap and objectives}
Vasculature biomarkers have been previously proposed and implemented in a semi or fully automated manner. However, these biomarkers were only analyzed individually across multiple groups. Hence the need to use a combined, comprehensive set of VBM within a machine learning (ML) framework for disease diagnosis and risk prediction. The large number of images needed to train ML algorithms require a fully automated computation of these VBM. In this paper we created a computerized toolbox, denoted PVBM, that enables the computation of 11 VBMs engineered from segmented arteriolar or venular networks. A potential application of PVBM is demonstrated by comparison of VBM in glaucoma patients versus healthy controls.

\section{Methods}

\subsection{Dataset}
A database provided by the University Hospitals of Leuven (UZ) in Belgium was used. This database contains 108,516 DFIs, centered around the optic disc. The resolution of $1444\times1444$ is higher than most public databases, which enables the visualization of smaller blood vessels. Median age was 66 (Q1 and Q3 respectively 54 and 75 years old) and $52\%$ were women. For a subset of the database, the blood vessels were manually segmented by retinal experts using Lirot.ai on Apple iPad Pro 11”and 13” \cite{FhimaLirot.ai:Segmentations}. This subset is denoted UZFG and consists of 69 DFIs. The patients included in UZFG were between 19 and 90 years old with a median age of 61 (Q1 and Q3 were 57 and 70 years old), $58\%$ were female. UZFG has $59\%$ of left and $41\%$ of right eye DFIs. Patients belonging to the UZFG are separated into the classes: Normal ophthalmic findings (NOR) and Glaucoma (GLA).



\begin{table*}[!tb]
\caption{Leuven A/V segmented database (UZFG) summary including the median (Q1-Q3) age, the gender and the diagnosis for the glaucoma (GLA) patient and normal ophthalmic findings (NOR) patient.}
\centering
\scalebox{0.9}{
    \begin{tabular}{c c c c}
    \hline\hline
    &N\textsuperscript{\underline{o}} DFIs & Age & Gender\\   
    \hline
    NOR&19&58(43-60)&42$\%$ M \\
    GLA&50&68(59-75)&42$\%$ M \\
    Total&69&61(57-70)&42$\%$ M \\
    \hline
    \end{tabular}
    }
\label{table:db}
\end{table*}


\subsubsection{Protocol for DFI vasculature segmentation by experts:} Arterioles carry higher concentration of oxyhemoglobin than venules and therefore exhibit a brighter inner part compared to their walls. This feature is known as the central reflex and is more typical for larger arterioles \cite{Miri2017AImages}. The exact intensity of this reflection is additionally influenced by the composition of the vessel wall \cite{Kaushik2007PrevalenceSign}, the roughness of the surface, the caliber of the blood column, the indices of refraction of erythrocytes and plasma, the pupil size, the axial length of the eye and the tilt of the blood vessel relative to the direction of incident light. Since the branching of arterioles and venules can be found inside the optic nerve head, two arterioles or venules can be found adjacently at the optic disc rim \cite{BrinchmannHansen1986TheoreticalVessels}. The image variability in term of color, contrast, and illumination, challenge an accurate (automatic) arteriole-venule discrimination \cite{Badawi2019MultilossVenules}. Vessel segmentation was manually performed by a pool of ten ophthalmology students experienced in microvascular research, using the software Lirot.ai \cite{FhimaLirot.ai:Segmentations} and afterwards corrected by an UZ retinal expert. Arteriole-venule discrimination was carried out based on the following visual and geometric features \cite{Miri2017AImages}:
\begin{itemize}
    \item Venules are darker than arterioles. 
    \item The central reflex is more recognizable in arterioles.
    \item Venules are usually thicker than arterioles.
    \item Venules and arterioles usually alternate near the optic disc.
    \item It is unlikely that arterioles cross arterioles or that venules cross venules.
\end{itemize}

\subsection{Digital vasculature biomarkers}
A total of 11 VBMs were implemented in PVBM  (Table \ref{table:2}). The biomarkers were computed separately for arterioles and venules.

\begin{table*}[!tb]
\caption{List of digital vasculature biomarker implemented in PVBM.}
\centering
\begin{tabular}{c c c c c c c c}
\hline\hline
Number&Biomarker & Definition & Unit \\ [0.5ex] 
\hline
1&OVLEN\cite{Martinez-Perez2000GeometricalImages} &Overall length&Pixel \\
2&OVPER\cite{Martinez-Perez2000GeometricalImages}&Overall perimeter&Pixel \\
3&OVAREA\cite{Martinez-Perez2000GeometricalImages}&Overall area&Pixel$^2$ \\
4&END\cite{Martinez-Perez2000GeometricalImages}&Number of endpoints &- \\
5&INTER\cite{Martinez-Perez2000GeometricalImages}&Number of intersection points &- \\
6&TOR\cite{Martinez-Perez2000GeometricalImages,Lotmar1979MeasurementPhotographs}&Median tortuosity&$\%$ \\
7&BA\cite{Martinez-Perez2000GeometricalImages}&Branching angles &$^o (degree)$ \\
8&$D_{0}$\cite{Stosic2006MultifractalVessels,Chhabra1989DirectSpectrum} & Capacity dimension &- \\
9&$D_{1}$\cite{Stosic2006MultifractalVessels,Chhabra1989DirectSpectrum} & Entropy dimension &- \\
10&$D_{2}$\cite{Stosic2006MultifractalVessels,Chhabra1989DirectSpectrum} & Correlation dimension &- \\
11&SL\cite{Macek2009EvolutionHeliosphere} & Singularity Length &- \\
\hline
\end{tabular}
\label{table:2}
\end{table*}

\subsubsection{Overall length:}

\begin{figure*}[!tb]
    \centering
	\includegraphics[width=0.6\textwidth]{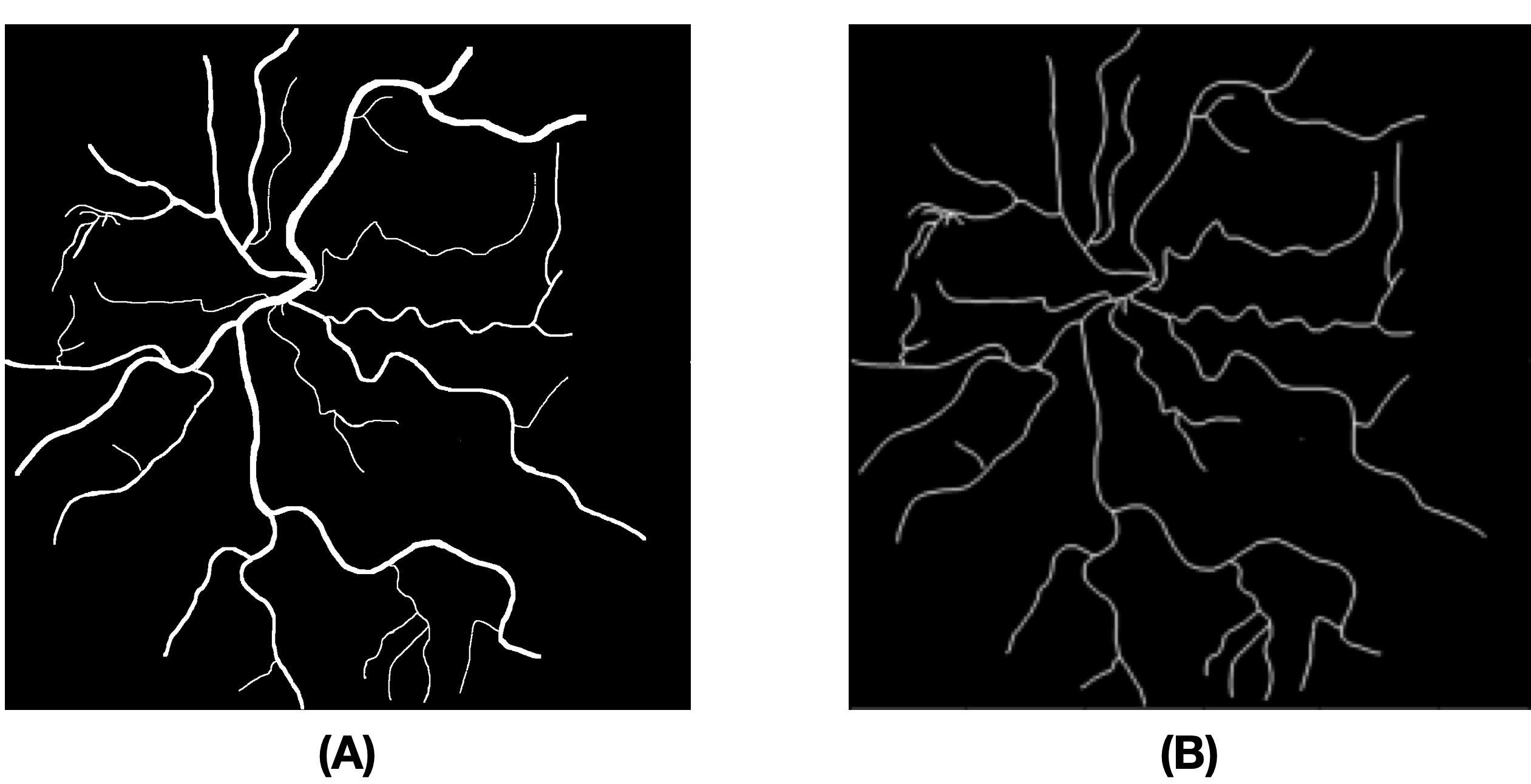} 
	\caption{Skeletonization process of a vascular network. (A): Example of a vascular network $y_{a}$, (B): Corresponding skeleton of $y_{a}$.} 
	\label{fig:skelet}
\end{figure*}

The OVLEN biomarker refers to the sum of the length of a vascular network, whether for arterioles or venules. To compute it, the first step is to extract the skeleton of the vascular network, which can be seen in fig. \ref{fig:skelet}. Then the number of pixels that belong to this skeleton are summed, and divided by the image size (1444x1444), then multiplied by 1000 for scaling purposes. It is computed using the following formula:
\begin{equation}
    OVLEN = 1e3*\frac{\Sigma_{p \in S} \sqrt{2}*\mathbb{1}_{|\partial_x p| + |\partial_y p| = 0}(p) + \mathbb{1}_{|\partial_x p| + |\partial_y p| \neq 0}(p)}{1444^2}
\end{equation}
\noindent where $S$ is the set of pixel inside the skeletonized image (Fig.\ref{fig:skelet}), and x/y represent the horizontal/vertical direction.

\begin{figure*}[!tb]
    \centering
	\includegraphics[width=0.6\textwidth]{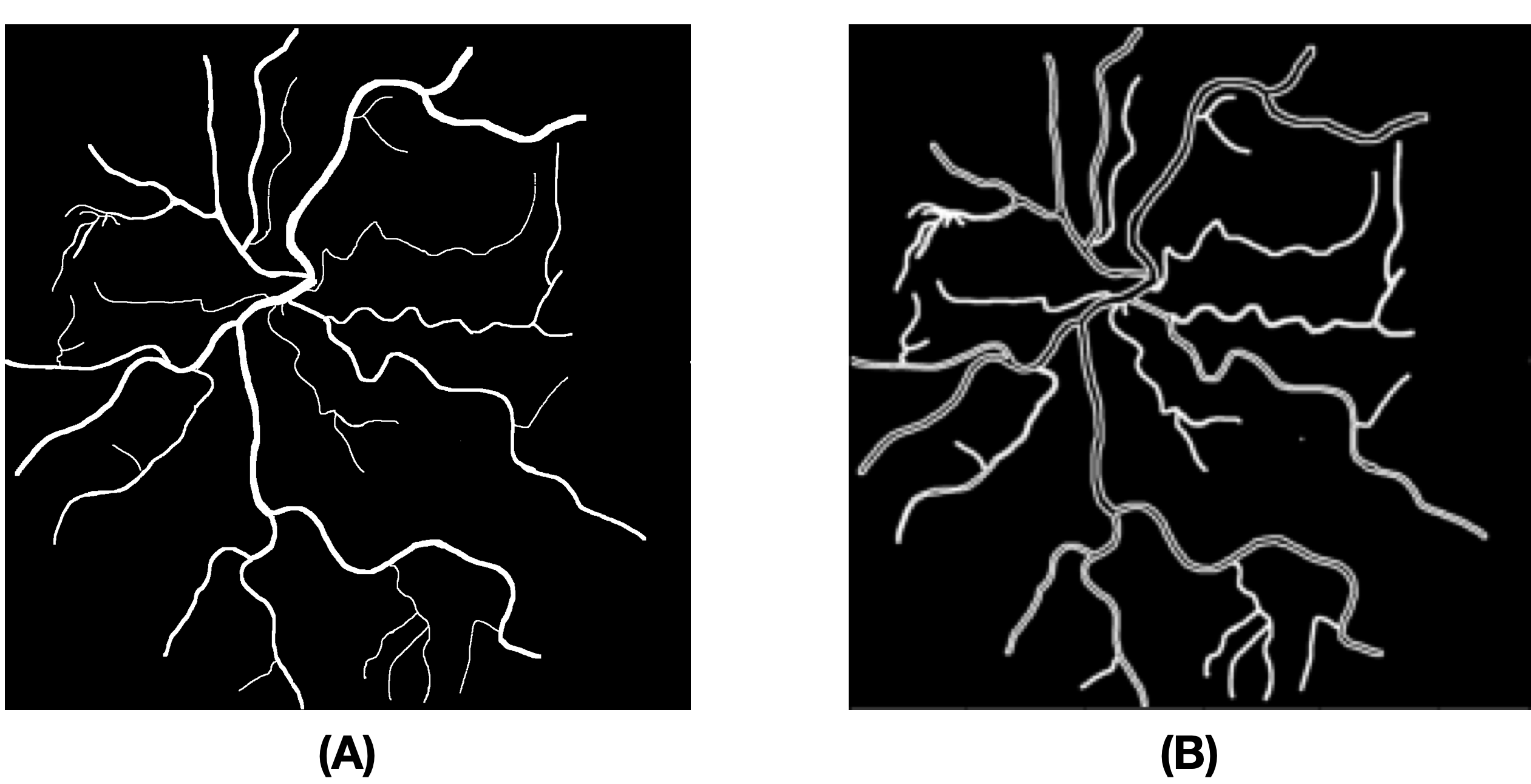} 
	\caption{Border computation process of a vascular network. (A): Vascular network $y_{a}$, (B): Corresponding computed edge of $y_{a}$.}
	\label{perim}
\end{figure*}

\subsubsection{Overall perimeter:}
The OVPER refers to the sum of perimeter's length of a vascular network. It is computed as the length of the border of the overall segmentation. It required an edge extraction of the segmentation, which could be easily found by convolving the original segmentation by a Laplacian filter (Fig. \ref{perim}). It is computed using the following formula:
\begin{equation}
    OVPER = 1e2*\frac{\Sigma_{p \in E} \sqrt{2}*\mathbb{1}_{|\partial_x p| + |\partial_y p| = 0}(p) + \mathbb{1}_{|\partial_x p| + |\partial_y p| \neq 0}(p)}{1444^2}
\end{equation}

\noindent where $E$ is set of pixel inside the edge image of the vascular network (Fig. \ref{perim}) and x/y represent the horizontal/vertical direction.

\subsubsection{Overall area: }
OVAREA is defined as the surface covered by the segmentation. In terms of pixels, it could be represented as the ratio of white pixels in the segmentation to the overall number of pixels. It is computed using the following formula:
\begin{equation}
OVAREA = 1e2*\frac{\Sigma_{p \in V} \mathbb{1}_{1}(p)}{1444^2}
\end{equation}
\noindent where $V$ is set of pixel inside the image of the vascular network (Fig. \ref{perim}).

\subsubsection{Endpoints and intersection points: }
\begin{figure*}[!tb]
    \centering
	\includegraphics[width=0.8\textwidth]{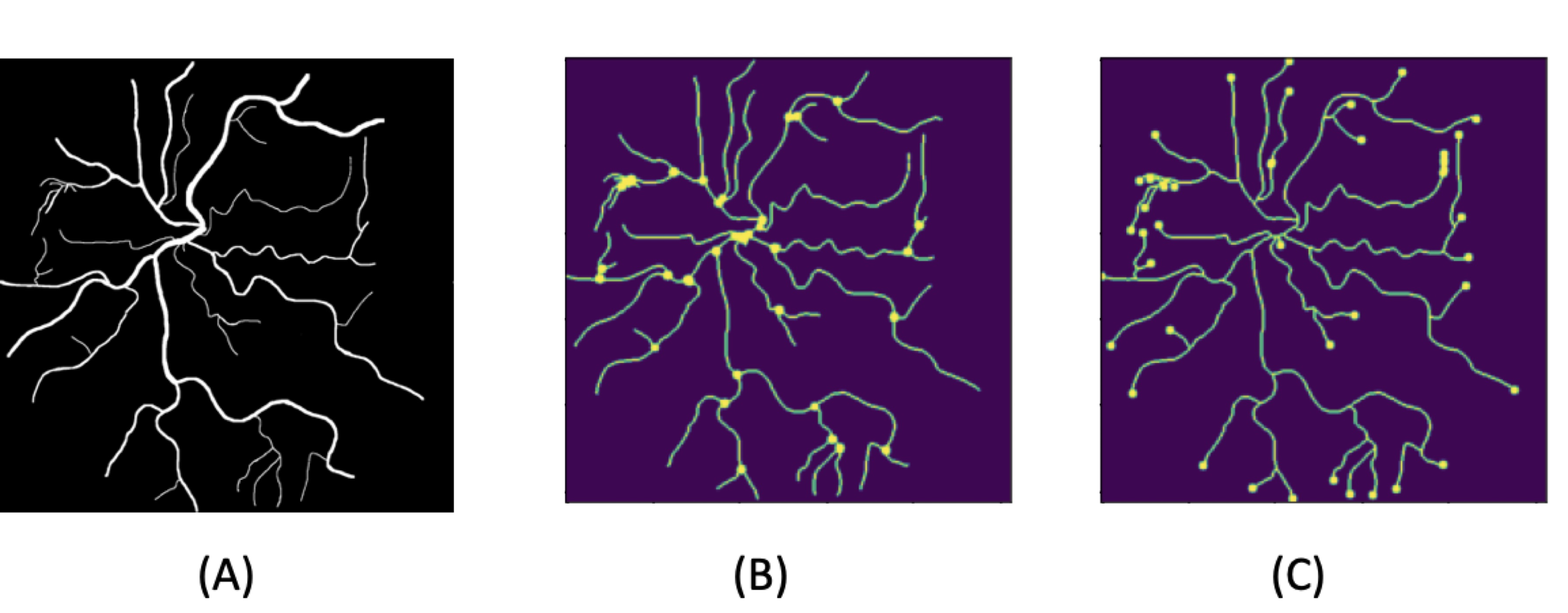} 
	\caption{Automatic detection of the particular points of a vascular network. (A): Vascular network $y_{a}$, (B): Automatic detection of the intersection points (INTER = 32) of $y_{a}$, (C): Automatic detection of the endpoints (END = 40) of $y_{a}$.}
	\label{particular}
\end{figure*}

The endpoints are the points at the end of the vascular network, which means in the skeleton version of the network, the points which have only one neighbor which belongs to the skeleton. The intersection points are the points where a blood vessel is divided into more than one blood vessel, which means in the skeleton version of the network, the points which have more than two neighbors which belong to the skeleton. Their automatic detection was done using a filter $k$ of size (3x3) where $k_{i,j} = 10$ if $i=j$, or $k_{i,j} = 1$ otherwise. The skeleton is then convolved with this filter to obtain a new image. In this new image, the endpoints will be the pixels with a value of 11, and the intersection points will be the pixels with a value of 13 or larger (Fig. \ref{particular}). We can represent the endpoints and the intersection points according to the following equation:
\begin{equation}
    END =  \{p = (i,j) \in Skeleton | (Skeleton \circledast 
    \begin{pmatrix}
      1 & 1 & 1\\
     1 & 10 & 1\\
     1&1&1
     \end{pmatrix})[i,j] = 11 \}
\end{equation}
\begin{equation}
    INTER =  \{p = (i,j) \in Skeleton | (Skeleton \circledast 
    \begin{pmatrix}
      1 & 1 & 1\\
     1 & 10 & 1\\
     1&1&1
     \end{pmatrix})[i,j] \geq 13 \}
\end{equation}

Particular points is the name given to the combinated set of points resulting from the union of the endpoints and intersection points. The number of endpoints (END) and the number of intersection points (INTER) were computed as VBMs.

\subsubsection{Median tortuosity: }
\begin{figure}[ht]
    \centering
	\includegraphics[width=0.6\textwidth]{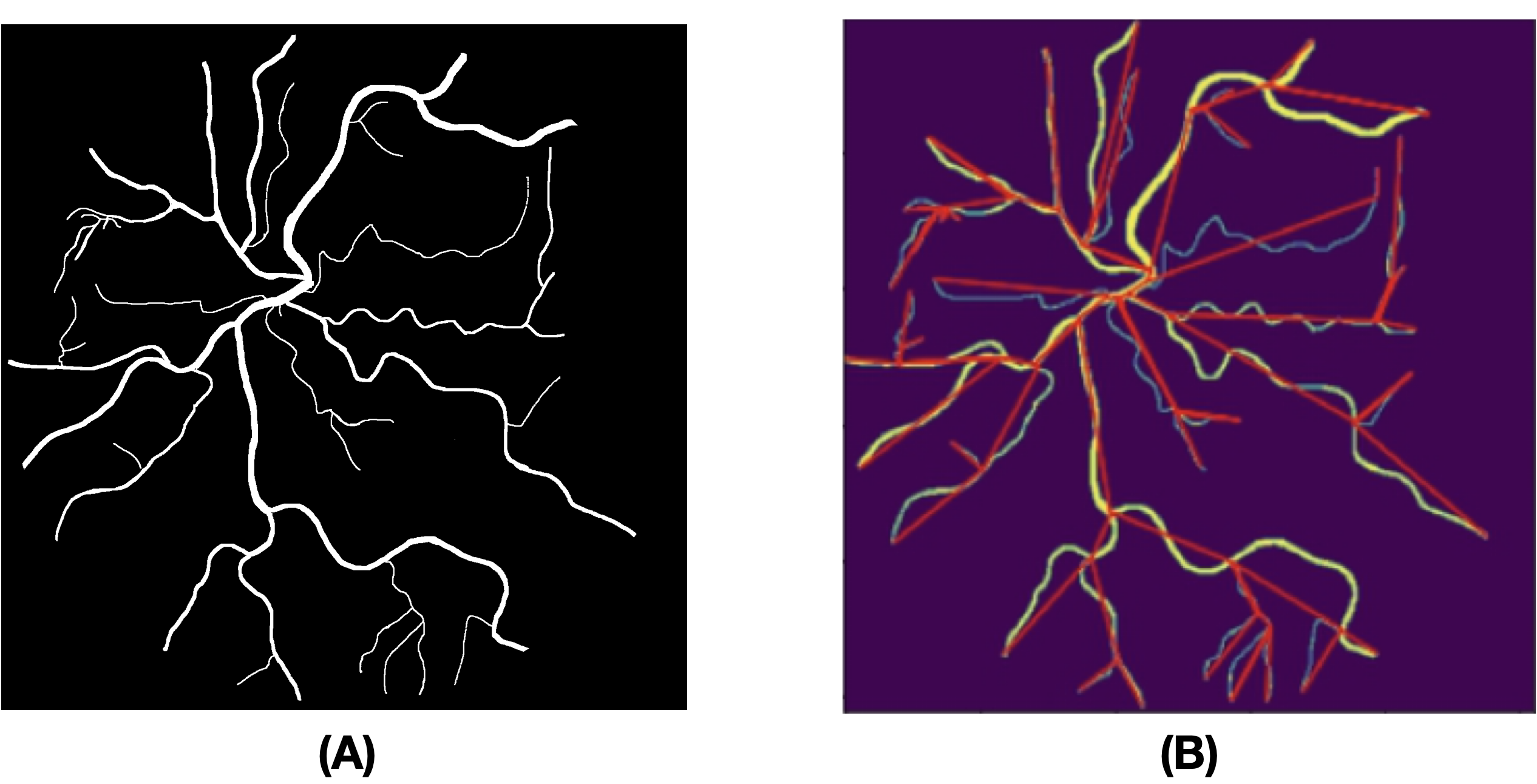} 
	\caption{Computing the linear interpolation of a vascular network. (A): Vascular network $y_{a}$ and (B): Linear interpolation between all the particular points of $y_{a}$.}
	\label{fig:secfig}
\end{figure}

The simplest mathematical method to estimate tortuosity is the arc-chord ratio, defined as the ratio between the length of the curve and the distance between its ends \cite{Hart1999MeasurementTortuosity}. In our work, the median tortuosity was computed using the arc-chord ratio based on the linear interpolation of all the blood vessels. For that purpose, the skeleton is treated as a graph and particular points are extracted. To compute the linear interpolation it is then required to find all the particular points connected to a given particular point. The connection between each particular point was stored in a dictionary according to Algorithm 1; the output of this algorithm is a dictionary where the keys are the particular points, and the values are the list of the connected particular points. Having this dictionary, it is possible to generate the linear interpolation between each connected particular point, as it is shown in Fig. \ref{fig:secfig}. The tortuosity of each blood vessel can be estimated by computing the ratio between the blood vessel's length (yellow curves in Fig. \ref{fig:secfig}) and the length of the interpolation of this blood vessel (red lines in Fig. \ref{fig:secfig}). The median tortuosity will then be the median value of the tortuosity of all the blood vessels.

\begin{verbatim}
Algorithm 1
program Compute connected pixel dictionary ( S (= skeleton),
P (= particular point list))
   initialize an empty dictionary: connected;
   initialize an empty dictionary: visited;
   for (i,j) in P:
       if S[i,j] == 1 (White):
          recursive(i,j,S,i,j,visited,P,connected);
end.

program recursive(i_or, j_or, S (= skeleton),
i,j,visited, P(= particular point list) ,connected )
   up = (i-1,j);
   down = (i+1,j);
   left = (i,j-1);
   right = (i,j+1);
   up left = (i-1,j-1);
   up right = (i-1,j+1);
   down left = (i+1,j-1);
   down right = (i+1,j+1);
   if up[0] >= 0 and visited.get(up,default = 0) == 0:
        if up not in P:
            visited[up] = 1;
        if S[up[0]][up[1]] == 1 (White):
        point = up;
        if point is in P:
            connected[i_or,j_or] = 
            connected.get((i_or,j_or),default = []) + [up];
        else:
            recursive(i_or, j_or,  S, point[0],
            point[1],visited,P ,connected);
     
    Do equivalent things for down, left, right, up left, up right,
    down left, down right.

end.

\end{verbatim}

\subsubsection{Branching angles:}
\begin{figure*}[!tb]
    \centering
	\includegraphics[width=0.7 \textwidth]{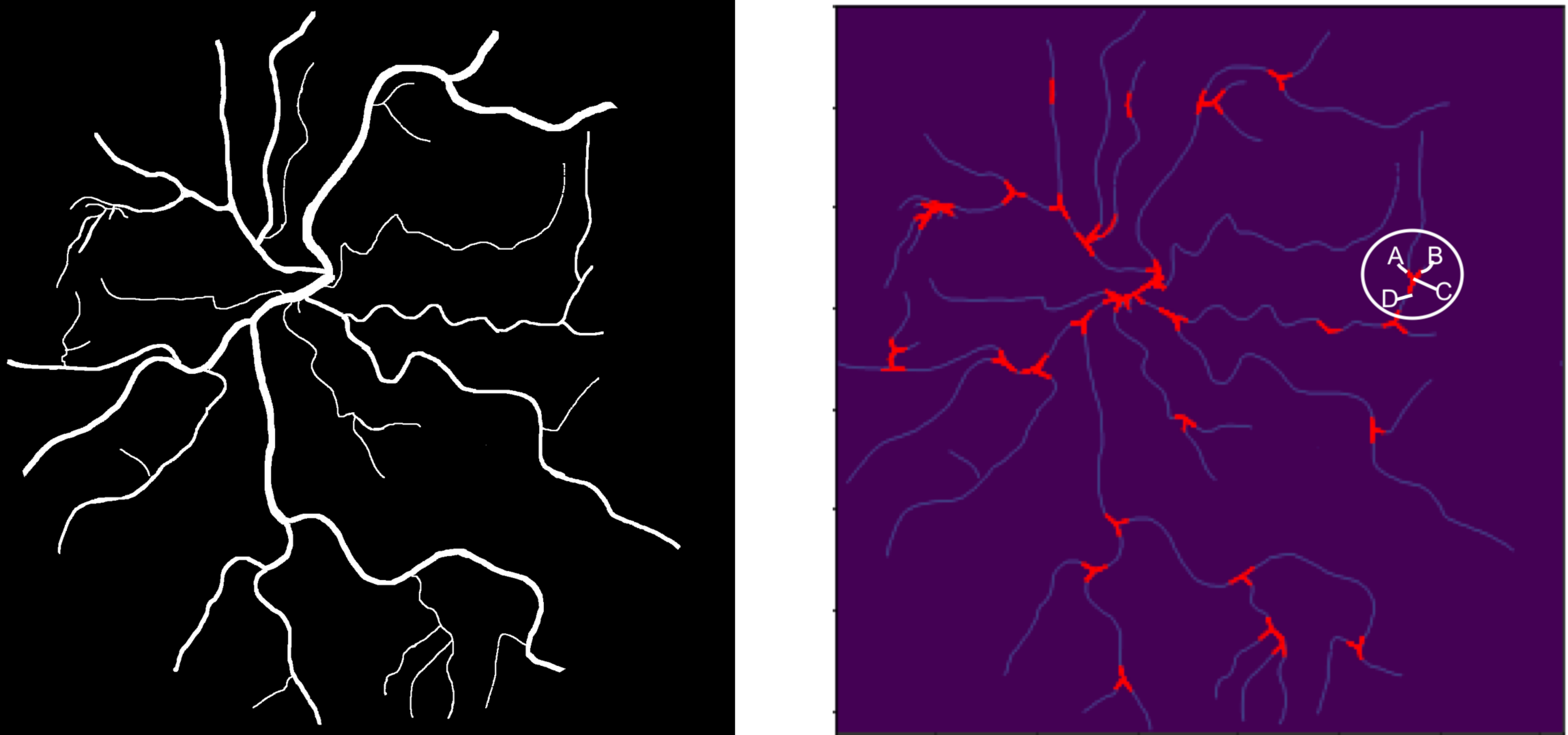} 
	\caption{Computation of all the angle of a vascular network.}
	\label{fig:ang}
\end{figure*}

A vascular network’s branching angles (BA) can be defined as the angle where a blood vessel is divided into smaller blood vessels. The computation of BA is performed using the following steps: starting by extracting all the angles of a vascular network using a simple modification of the linear interpolation algorithm that we developed to compute the tortuosity (Fig. \ref{fig:ang}). To extract only the branching angles, all other angles need to be discarded. For instance, in Fig. \ref{fig:ang} an ellipse has been drawn around a branching angle, giving us the following four points $A$, $B$, $C$ and $D$. These points define the following two segments: [$A$, $C$] and [$B$, $D$] with their intersection point $C$. Three different angles can be computed from this branching: $\widehat{ACD}$, $\widehat{ACB}$ and $\widehat{BCD}$. The branching angle corresponds to $\widehat{ACB}$. We need to define the centroid of the graph in order to find the only relevant angle.
\\
In a connected graph, we define a centroid as the closest point to any other particular point of the graph. To compute the centroid, we will need to extract a set of points $S$, which will be our particular points in a connected graph;

\begin{equation}
S = \{p, p \in skeleton, N(p) \not\in \{2,0\}\}
\end{equation}
where $N(p)$ is the number of neighbouring pixels of $p$ which belong to the skeleton.

\begin{figure*}[!tb]
    \centering
	\includegraphics[width=0.8 \textwidth]{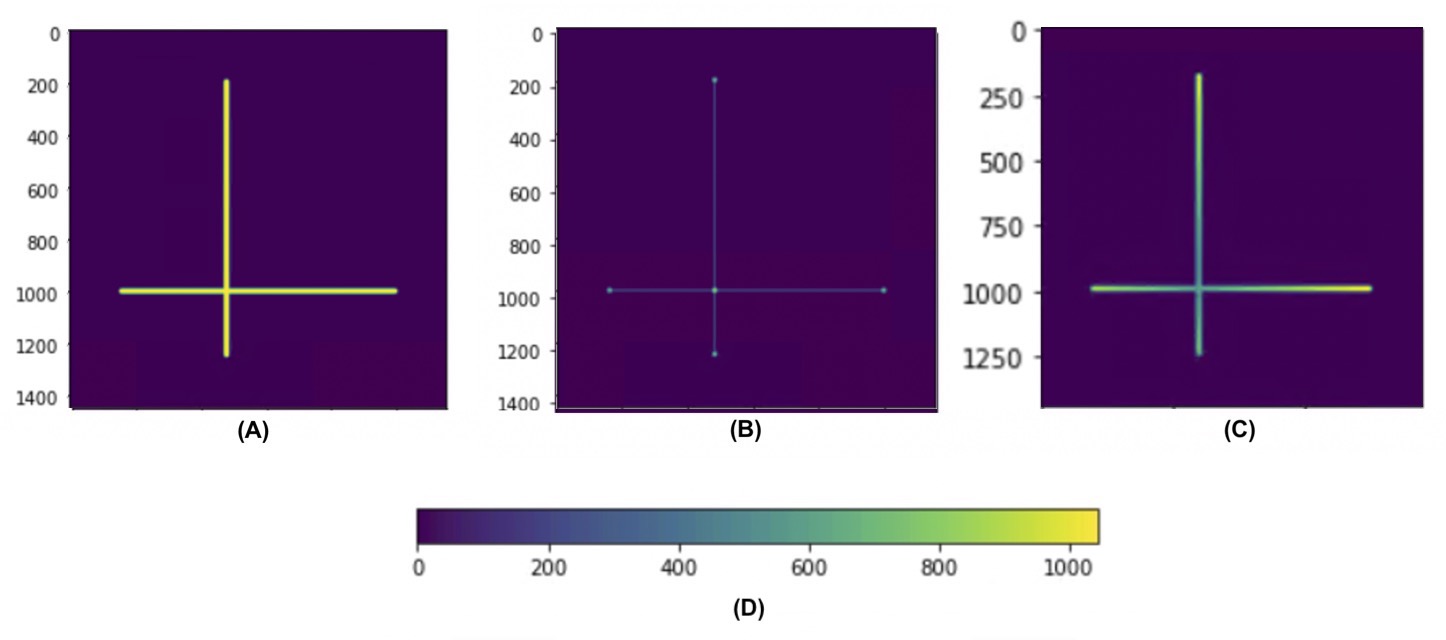} 
	\caption{Computation of the centroid of a simple graph. (A): Original graph, (B): Particular point detection, (C) Value of the $f(.)$ function for each pixel of this graph, (D): Heatmap where the centroid is the point with the lowest value according to the function $f(.)$.}
	\label{fig:ang2}
\end{figure*}

\begin{figure*}[!tb]
    \centering
	\includegraphics[width=0.7 \textwidth]{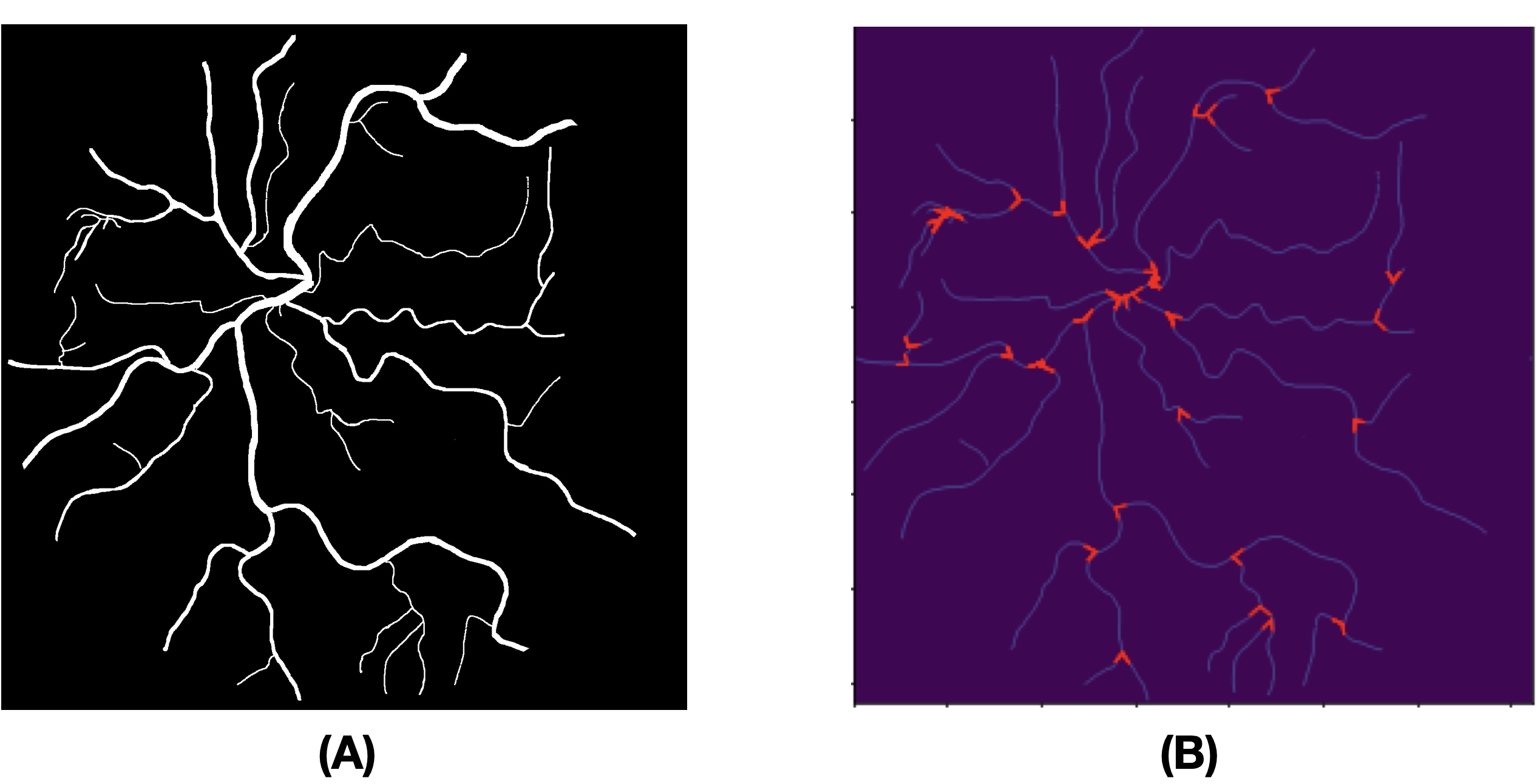} 
	\caption{Automated computation of the branching angles of the vascular network. (A): Original segmentation, (B): Branching angles detected automatically.}
	\label{fig:ang3}
\end{figure*}

Then we create a metric $f$ such that for any point $p$ in the skeleton of the segmentation: 

\begin{equation}
f(p)=max_{s \in skeleton} dist(p,s)
\end{equation}

\noindent where $dist$ is the distance, measured as the number of pixels required to reach $s$ from $p$ by staying inside the skeleton. The centroid will naturally be the point with the lowest value according to this function. And for a random point $p$, the higher the value of $f(p)$, the farther $p$ is from the centroid. A simple example can be seen in Fig. \ref{fig:ang2}. This also generalizes to segmentation with multiple disconnected parts, assuming that each part has a centroid. 
The branching angle between $A$, $B$, $C$, and $D$ is the angle between the 3 points that are the farthest from the centroid of the blood vessel in terms of pixel distance when you navigate through the graph of the vascular network which is equivalent to our challenge of deleting the closest point to the centroid of the blood vessel. It is possible to delete the irrelevant point thanks to this centroid detection, and to compute the set of the branching angles  $\Gamma = \{BA_{i}\}_{i=1:n}$ automatically (Figure \ref{fig:ang3}). The BA biomarker is defined by the median of all the found angles.
\begin{equation}
BA = median\{a \in \Gamma\} 
\end{equation}
\noindent where $\Gamma$ is the set of the detected branching angles.

\subsubsection{Fractal dimensions:}
The retinal blood vessels form a complex branching pattern that can be quantified using fractal dimension (monofractal) \cite{Mandelbrot1982TheNature,Mainster1990TheImplications,Family1989FractalVessels}. In fractal geometry, the fractal dimension is a measure of the space-filling capacity of a pattern that tells how a fractal scales differently from the space it is embedded in. Fractal dimension can be thought of as an extension of the familiar Euclidean dimensions allowing intermediate states. The fractal dimension of the retinal vascular tree lies between 1 and 2, indicating that its branching pattern fills space more thoroughly than a line, but less than a plane \cite{Mainster1990TheImplications}. The fractal dimension measures the global branching complexity, which can be altered by the vessel rarefaction, proliferation and other anatomical changes in a pathological scenario. The simplest and most common method used in the literature for monofractal calculation is the box-counting method \cite{Mandelbrot1982TheNature}.

However, a single monofractal is limited in describing human eye retinal vasculature. It has been observed that retinal vasculature has multifractal properties which are a generalized notion of a fractal dimension \cite{Stosic2006MultifractalVessels,Talu2013CharacterizationDescriptors,Talu2017AnalysisGeometry}. Multifractal dimensions are characterized by a hierarchy of exponents which can reveal more complex geometrical properties in a structure \cite{Talu2013CharacterizationDescriptors}. The most common multifractal dimensions for measuring retinal vasculature are $D_{0},D_{1},D_{2}$ which satisfy the inequality $D_{0}>D_{1}>D_{2}$ \cite{Stosic2006MultifractalVessels,Posadas2001MultifractalDistributions} and also called the capacity dimension (a monofractal), entropy dimension and correlation dimension respectively. Following \cite{VanCraenendonck2021RetinalPopulation}, we have implemented $D_{0},D_{1},D_{2}$ and Singularity length in a similar manner to Chhabra et al. \cite{Chhabra1989DirectSpectrum} and the commonly used plugin FracLac \cite{Karperien2013FraclacImagej} for ImageJ software \cite{Rasband2011ImagejUsa}. The generalized multifractal dimensions are defined as:

\begin{equation}\label{dq}
D_{q}=\begin{cases}
-\lim_{\epsilon\rightarrow1}\frac{\sum_{i}P_{i}\left(\epsilon\right)\log P_{i}\left(\epsilon\right)}{\log\epsilon} & ,q=1\\\\
\frac{1}{q-1}\lim_{\epsilon\rightarrow0}\frac{\log\sum_{i}P_{i}^{q}\left(\epsilon\right)}{\log\epsilon} & o.w.
\end{cases}
\end{equation}
where $P_{i}(\epsilon)$ is the pixel probability in the $i^{\text{th}}$ grid box sized $\epsilon$, and $q$ is the order of the moment of the measure. In addition, multifractals can be described by a singularity spectrum $f\left(\alpha\right)-\alpha$ which is related to the $D_{q}-q$ spectrum by the Legendre transformation. In order to calculate $f\left(\alpha\right)-\alpha$ spectrum, we use an alternative approach described by \cite{Chhabra1989DirectSpectrum}:
\begin{equation}\label{fq}
f\left(q\right)	=\lim_{\epsilon\rightarrow0}\frac{\sum_{i}\mu_{i}\left(q,\epsilon\right)\log\mu_{i}\left(q,\epsilon\right)}{\log\epsilon}
\end{equation}
\begin{equation}\label{aq}
\alpha\left(q\right)	=\lim_{\epsilon\rightarrow0}\frac{\sum_{i}\mu_{i}\left(q,\epsilon\right)\log P_{i}\left(\epsilon\right)}{\log\epsilon}
\end{equation}
\begin{equation}
\mu_{i}\left(q,\epsilon\right)	=\frac{P_{i}\left(\epsilon\right)^{q}}{\sum_{i}P_{i}\left(\epsilon\right)^{q}}    
\end{equation}
The $f\left(\alpha\right)-\alpha$ spectrum is characterized by a bell shaped curve with one maxima point. From this curve, an additional biomarker is computed - the spectrum range $\Delta\alpha$ which is also called Singularity Length (SL).
\begin{equation}
    \alpha_{\text{min}}=\lim_{q\rightarrow\infty}\alpha\left(q\right),\:\alpha_{\text{max}}=\lim_{q\rightarrow-\infty}\alpha\left(q\right)
\end{equation}
\begin{equation}
    \Delta\alpha=\alpha_{\text{max}}-\alpha_{\text{min}}
\end{equation}
SL quantifies the multifractality degree \cite{Macek2009EvolutionHeliosphere} of an image.

 
The calculation of equations \ref{dq}, \ref{fq}, \ref{aq} was done by linear regression with a linear set of box sizes $\epsilon$ for every $q$. The values of these graphs are sensitive to grid placement on the segmented DFI image, as they depend on pixel distribution across the image. We followed a similar optimization method as used by the FracLac plugin \cite{Karperien2013FraclacImagej}, which is to change the sampling grid location by rotating the image randomly, and to choose the measurement which satisfies the inequality $D_{0}>D_{1}>D_{2}$ and has the highest value of $D_{0}$. In addition, to overcome numerical issues saturated grid boxes and grid boxes with low occupancy of pixels were ignored. $\alpha_{\text{min}}$ and $\alpha_{\text{max}}$ were estimated by $\alpha_{\text{min}}\approx\alpha\left(q=10\right)$ and $\alpha_{\text{max}}\approx\alpha\left(q=-10\right)$.

\subsubsection{Benchmark against existing software:}
In order to validate the implementation of some of the biomarkers we implemented in PVBM we benchmarked their values against two ImageJ plugins, namely AnalyzeSkeleton \cite{ArgandaCarreras20103DTissue} and FracLac \cite{Karperien2013FraclacImagej}. The benchmark was performed for the arterioles from the entire UZFG dataset. A direct benchmark could be performed for the following biomarkers: OVAREA, TOR, D\textsubscript{0}, D\textsubscript{1}, D\textsubscript{2} and SL. An extrapolated comparison could be performed for the OVLEN. Indeed, these were not directly outputted by the plugin, but could be derived from the AnalyzeSkeleton \cite{ArgandaCarreras20103DTissue} plugin.  No benchmark could be performed for OVPER, END, INTER and BA because of the lack of open source available benchmark software.

\section{Results}

Table $\ref{table:3}$ shows that the VBMs benchmarked against reference software had  very close values with normalized root mean square errors ranging from 0 to 0.316.

\begin{figure*}[!bt]
    \centering
	\includegraphics[width=0.7\textwidth]{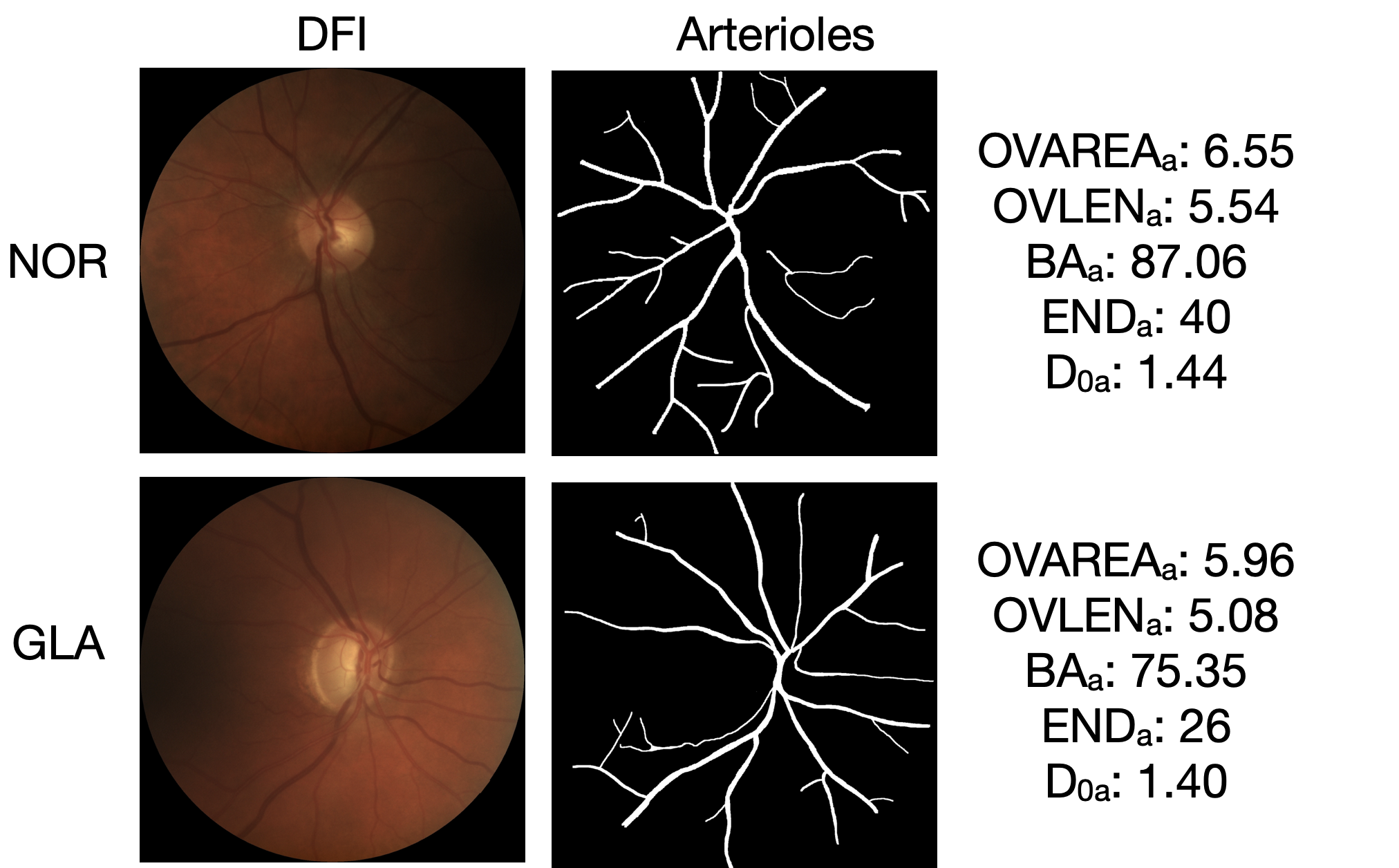}
	\caption{Example of biomarkers computed from the arterioles of two DFIs. The first row is a DFI of a healthy control (NOR) and the second column is from a glaucoma patient (GLA). The OVAREA biomarker shows a larger vascular area in the NOR images. The BA biomarker shows that the branching of the NOR image are bigger than the one of the GLA image. Finally, the END biomarker indicates that the GLA image had less arteriolar branching compared to healthy controls, leading to a lower number of endpoints.}
	\label{exe}
\end{figure*}

Table $\ref{table:biom1}$ and $\ref{table:biom}$ provide summary statistics and a statistical analysis of the VBMs for the GLA NOR groups. The statistics were presented as median and interquartile (Q1 and Q3), and the p-value from the Wilcoxon signed-rank. The arteriolar OVAREA, OVLEN, and END were the most significant in distinguishing between the two groups. Fig. \ref{exe} presents qualitative examples of three DFIs with arteriolar OVAREA, OVLEN, BA, END and D\textsubscript{0} VBM values.


\begin{table*}[!b]
\caption{PVBM benchmark against reference ImageJ plugins using the arterioles of the UZFG dataset. $\mu$: mean, $\sigma$: standard deviation, RMSE: root mean square error, NRMSE: normalized root mean square error.}
\centering
\scalebox{0.9}{
\begin{tabular}{l l c c c c c c c}
\hline\hline
Biomarker&Benchmark&\multicolumn{2}{c}{Benchmark} & \multicolumn{2}{c}{This work} & \multicolumn{2}{c}{Difference}\\
&&\multicolumn{2}{c}{results} & \multicolumn{2}{c}{} & \multicolumn{2}{c}{} \\ [0.5ex] %
\hline
& &$\mu$ & $\sigma$ & $\mu$ & $\sigma$ & RMSE & NRMSE\\ [0.5ex] 
\hline
OVLEN&AnalyzeSkeleton \cite{ArgandaCarreras20103DTissue}&4.687 &0.928 & 4.868 &0.974& 0.194& 0.060& \\
OVAREA&ImageJ \cite{Rasband2011ImagejUsa}&4.939 &0.011 & 4.939 &0.011& 0& 0& \\
TOR&AnalyzeSkeleton \cite{ArgandaCarreras20103DTissue}&1.081&0.007&1.084&0.007& 0.003& 0.11 \\
$D_{0}$&FracLac \cite{Karperien2013FraclacImagej}&1.373 &0.035 & 1.425 &0.026& 0.054& 0.301& \\
$D_{1}$&FracLac \cite{Karperien2013FraclacImagej}&1.367 &0.034 & 1.390 &0.027& 0.028& 0.155& \\
$D_{2}$&FracLac \cite{Karperien2013FraclacImagej}&1.361  &0.033 & 1.375 &0.028& 0.021& 0.115& \\
$SL$&FracLac \cite{Karperien2013FraclacImagej}   &0.626  &0.104  & 0.645 &0.076& 0.128& 0.316& \\
\hline
\end{tabular}
}
\label{table:3}
\end{table*}

\begin{table*}[!t]
\caption{Summary of the biomarkers analyzed for arterioles (\textsubscript{a}) and extracted with the PVBM toolbox with their median (Q1-Q3). Refer to Table $\ref{table:2}$ for the definition of the VBM acronyms. P-values are provided using the Wilcoxon signed-rank test.}
\centering

\scalebox{1}{
    
    \begin{tabular}{c c c c c}
    
\hline\hline
 &NOR (n=19) & GLA (n=50) & p\\ [0.5ex] 
\hline
OVAREA\textsubscript{a}&5.82 (5.26-6.23)&4.52 (3.76-5.44)& $\textbf{4e-4}$\\
\hline
OVLEN\textsubscript{a}&5.53 (5.28-6.21)&4.85 (4.33-5.37)& $\textbf{1e-4}$\\
\hline
OVPER\textsubscript{a}&3.94 (3.74-4.33)&3.44 (3.01-3.80)& $\textbf{2e-4}$\\
\hline
BA\textsubscript{a}&87.29 (81.88-93.28)&81.93 (75.41-87.87)&$\textbf{2e-2}$\\
\hline
END\textsubscript{a}&37.0 (28.5-43.5)&27.0 (21.0-32.75)&$\textbf{7e-4}$\\
\hline
INTER\textsubscript{a}&36.0 (28.5-43.0)&24.5 (20.0-30.0)&$\textbf{8e-4}$\\
\hline
TOR\textsubscript{a}&
1.08 (1.08-1.09)&
1.08(1.08-1.09)&$1e-1$\\
\hline
D\textsubscript{0a}&1.44 (1.42-1.45)&1.42 (1.40-1.44)&$\textbf{5e-3}$\\
\hline
D\textsubscript{1a}&1.41 (1.38-1.42)&1.38 (1.37-1.40)&$\textbf{3e-3}$\\
\hline
D\textsubscript{2a}&1.39 (1.37-1.40)&1.37 (1.35-1.38)&$\textbf{4e-3}$\\
\hline
SL\textsubscript{a}&0.62 (0.59-0.64)&0.64 (0.62-0.78)&$\textbf{9e-3}$\\
\hline
\end{tabular}
}
\label{table:biom1}
\end{table*}

\begin{table*}[!t]
\caption{Summary of the biomarkers analyzed for venules (\textsubscript{v}) and extracted with the PVBM toolbox with their median (Q1-Q3). Refer to Table $\ref{table:2}$ for the definition of the VBM acronyms. P-values are provided using the Wilcoxon signed-rank test.}
\centering

\scalebox{1}{
    
    \begin{tabular}{c c c c c}
    
\hline\hline
 &NOR (n=19) & GLA (n=50) & p\\ [0.5ex] 
\hline

OVAREA\textsubscript{v}&6.3 (5.82-6.71)&5.28 (4.59-6.01)&$\textbf{9e-4}$\\
\hline
OVLEN\textsubscript{v}&5.42 (4.82-5.67)&4.69 (4.21-5.13)&$\textbf{1e-3}$\\
\hline
OVPER\textsubscript{v}&3.77 (3.36-3.99)&3.26 (2.96-3.59)&$\textbf{3e-3}$\\
\hline
BA\textsubscript{v}&84.14 (78.74-86.9)&85.1 (81.52-91.08)&$1e-1$\\
\hline
END\textsubscript{v}&339.0 (31.5-43.0)&28.5 (23.0-34.0)&$\textbf{1e-3}$\\
\hline
INTER\textsubscript{v}&36.0 (31.5-42.5)&27.0 (21.0-33.0)&$\textbf{1e-3}$\\
\hline
TOR\textsubscript{v}&1.08 (1.08-1.09)&1.08
 (1.08-1.09)& $4e-1$\\
\hline
D\textsubscript{0v}&1.43 (1.4-1.44)&1.40 (1.38-1.42)& $\textbf{4e-3}$\\
\hline
D\textsubscript{1v}&1.38 (1.36-1.40)&1.37 (1.35-1.38)& $\textbf{8e-3}$\\
\hline
D\textsubscript{2v}&1.36 (1.34-1.39)&1.35 (1.33-1.36)& $\textbf{1e-2}$\\
\hline
SL\textsubscript{v}&0.61 (0.58-0.66)&0.63 (0.59-0.66)& $2e-1$\\
\hline
\end{tabular}
}
\label{table:biom}
\end{table*}

\section{Discussion and future work}
The first contribution of this work is the creation of a toolbox for VBMs, which is made open source under a GNU GPL 3 license and will be made available on physiozoo.com (following publication). In particular, novel algorithms were introduced to estimate the tortuosity and branching angles.

The second contribution of this work is the application of the PVBM toolbox to a new dataset of manually segmented vessels from DFIs. The statistical analysis that we have performed showed that the arterioles-based biomarkers are the most significant in distinguishing between NOR and GLA. For arterioles and venules, all biomarkers were significant and lower in glaucoma patients compared to healthy controls except for tortuosity, venular singularity length and venular branching angles.

A limitation of our experiment is that although the images were taken with the same procedure, which includes the disk being centered, there is some variation in the exact location of the disk due to the non-automated operation. In future work, we need to consider the detection of the disk to delineate a circular frame centered on the disk to engineer the vasculature biomarkers more consistently. Furthermore, other biomarkers may be implemented such as the vessel diameter \cite{Goldenberg2013DiametersTomography}, CRAE \cite{Parr1974GeneralArtery}, CRVE \cite{Parr1974GeneralArtery}, branching coefficients \cite{Doubal2010RetinalImaging} which is the ratio of the sum of the cross-sectional areas of the two daughter vessels to the cross-sectional area of the parent vessel at an arteriolar bifurcation \cite{Doubal2010RetinalImaging}. Finally, it is to be studied to what extent VBMs may be evaluated from vasculature obtained using an automated (versus manual) segmentation algorithm as well as the effect of DFI quality \cite{AbramovichFundusQ-Net:Grading} on the results.


\clearpage
%
%
\bibliographystyle{splncs04}
\bibliography{references.bib}
\end{document}